\documentclass[runningheads]{llncs}
\usepackage{graphicx}
\usepackage{comment}
\usepackage{amsmath,amssymb} 
\usepackage{color}
\usepackage{mathtools}
\usepackage{pifont}

\usepackage{caption}
\usepackage{makecell}

\newcommand{\red}{\textcolor{black}}

\newcommand{\astrid}{\textcolor{black}}

\begin{document}
\pagestyle{headings}
\mainmatter
\def\ECCVSubNumber{02}  

\title{CLAWS: Clustering Assisted Weakly Supervised\\Learning with Normalcy Suppression for Anomalous Event Detection} 



\titlerunning{CLAWS Net}

\titlerunning{CLAWS Net for Anomalous Event Detection}
\author{Muhammad Zaigham Zaheer\inst{1,2} \and
Arif Mahmood\inst{3} \and
Marcella Astrid\inst{1,2} \and
Seung-Ik Lee\inst{1,2}}
\authorrunning{M. Z. Zaheer et al.}
%
\institute{University of Science and Technology, Daejeon, South Korea. \and
Electronics and Telecommunication Research Institute, Daejeon, South Korea.
\email{\{mzz, marcella.astrid\}@ust.ac.kr, the\_silee@etri.re.kr}\\
 \and
Information Technology University, Ferozpur Road, Lahore, Pakistan\\
\email{arif.mahmood@itu.edu.pk}}
\maketitle

\begin{abstract}

Learning to detect real-world anomalous events through video-level labels is a challenging task due to the rare occurrence of anomalies as well as noise in the labels. 
In this work, we propose a weakly supervised anomaly detection method which has manifold contributions including \red{1)} a random batch based training procedure to reduce inter-batch correlation, \red{2)} a normalcy suppression mechanism to minimize anomaly scores of the normal regions of a video by taking into account the overall information available in one training batch, and \red{3)} a clustering distance based loss to
contribute towards mitigating the label noise and to produce better anomaly representations by encouraging our model to generate distinct normal and anomalous clusters. The proposed method obtains 83.03\% and 89.67\% frame-level AUC performance on the UCF-Crime and ShanghaiTech datasets respectively, demonstrating its superiority over the existing state-of-the-art algorithms.
\keywords{Weakly Supervised Learning, Anomaly Detection, Abnormal Event Detection, Noisy Labeled Training, Event Localization}
\end{abstract}

\section{Introduction}

Anomalous event detection is an important computer vision domain because of its real-world applications in autonomous surveillance systems \astrid{\cite{mohammadi2016angry,sultani2010abnormal,kamijo2000traffic,sultani2018real,shanghaiTech2017}}. Attributed to the rare occurrences, anomalies are often seen as deviations from normal patterns, activities, or appearances. Hence, a widely popular approach for anomaly detection is to train a one-class classifier which can encode frequently occurring behaviors using only normal training examples \astrid{\cite{liu2018future,zhang2016video,luo2017revisit,xia2015learning,hinami2017joint,sabokrou2017deep_novelty,smeureanu2017deep,ravanbakhsh2018plug,ravanbakhsh2017abnormal,sultani2018real,hasan2016anomaly}}. Anomalies are then detected based on their distinction from the learned normalities. A drawback of such methods is the lack of representative training data capturing all variations of the normal behavior \astrid{\cite{chandola2009anomaly}}. 
Therefore, a new occurrence of a normal event may deviate enough from the trained patterns to be flagged as anomaly, hence causing false alarms \astrid{\cite{hasan2016anomaly}}. With the recent popularity in weakly supervised learning algorithms \astrid{\cite{liu2019completeness,liu2019weakly,yu2019temporal,narayan20193c,shou2018autoloc,wang2017untrimmednets}}, another interesting approach is to train a binary classifier using weakly labeled training data containing both normal and anomalous videos \astrid{\cite{sultani2018real,zhong2019graph}}.
A video is labeled as normal if all of its frames are normal and labeled as anomalous if some frames are abnormal. It means that a video labeled as anomalous may also contain numerous normal scenes. In such weakly supervised algorithms, neither temporal nor spatial annotations are needed which considerably reduces the laborious efforts required to obtain the fine-grained manual annotations of the training dataset.

Weakly supervised anomaly detection problem has recently been formulated as Multiple Instance Learning (MIL) task \cite{andrews2003support,sultani2018real}. A video is considered as a bag of segments in which each segment contains several consecutive frames and the training is carried out using video-level annotations by computing a ranking loss between the two top-scoring segments, one from normal and the other from anomalous bag \cite{sultani2018real}.
However, this method requires each video to have the same number of segments throughout the dataset which may not always be a practical approach. 
Since real-world scenarios may contain significantly varying length of videos, the events occurring in a small temporal range will not be represented well due to this rigid formulation.
More recently, another approach towards weakly supervised anomaly detection has also been proposed as learning under noisy labels in which the noise refers to normal segments within anomalous videos \cite{zhong2019graph}. Although it demonstrates superior performance, the method is
prone to data correlation since it is trained using a whole video at each training iteration. Such correlation can get even stronger in the case of datasets
recorded using stationary cameras. As reported in several existing works \astrid{\cite{bengio2012practical,mnih-atari-2013,mnih2015human}}, data correlation can significantly deteriorate the learning performance of a deep network. 

In contrast, we propose a batch based training architecture where a batch consists of several temporally consecutive segments of a video. Depending on the length of a video, several batches may be extracted from it.
In each training iteration, we arbitrarily select a batch across the whole training dataset to make the batches independent and identically distributed, thus eradicating inter-batch correlation for a stable and enhanced training. We still, however, utilize the temporal consistency information within a batch for better classification.
Extensive experiments demonstrated the efficacy of our proposed random batch selection mechanism as it yields significant performance improvements. Detailed discussion on this is provided in the results section.

Together with this batch based training scheme, a complementary attention-like mechanism may also contribute towards the improvement of an anomaly detection system. Various forms of attention have been introduced in many machine learning architectures with a common goal to highlight important regions within an input image or video
\astrid{\cite{chen2018attention,shen2018egocentric,woo2018cbam,hu2018squeeze,wang2017residual}}. Such mechanisms can be suitable with fully-supervised algorithms in which the attention layers is trained to highlight the important features corresponding to the class annotations of the training data. However, as our proposed approach is weakly supervised, we consider the problem as suppressing the features that correspond to normal events. 
Therefore, we propose a normalcy suppression mechanism that operates over a full batch and learns to suppress the normal features towards smaller values.
\red{Our formulation exploits the fact that the labels in normal training videos are noise free.}
Thus, in the case of an input containing anomalous portions, suppression tries to reduce the impact of normal regions within that input while keeping only the high anomaly regions active. Whereas, in the case of an input containing only normal portions, suppression distributes across the whole batch hence complementing the backbone network towards generating lower anomaly scores.

Furthermore, we also propose a clustering distance based loss, the intuition of which
is derived from the semi-supervised usage of clustering techniques \astrid{\cite{kamnitsas2018semi,fogel2019clustering,shukla2018semi}}. To this end, we propose unsupervised clustering to collaborate with our network resulting in an improved overall performance. We first assume two clusters considering that anomaly detection is a binary problem and an anomalous labeled video may also contain normal segments. Then, with each training iteration, the network attempts to minimize (or maximize) the inter-cluster distance in the case of a normal (or an abnormal) video. The clustering is performed on an intermediate representation taken from our backbone network. Therefore, minimization of our formulated clustering loss encourages the network to produce discriminative intermediate representations thus enhancing its anomaly detection performance.

The main contributions of this work are as follows:

\begin{itemize}
  \item Our CLAWS (CLustering Assisted Weakly Supervised) Net framework is trained in a weakly supervised fashion using only video-level labels to localize anomalous events.
  \item We propose a simple yet effective random batch selection scheme which enhances the performance of our system by removing inter-batch correlation.
 \item We also propose a normalcy suppression mechanism which, by exploiting full information of a batch, learns to suppress the features corresponding to the normal portions of an input.
 \item We formulate a clustering distance based loss which encourages the network to decrease the distance between the clusters created using a normal video and increase it for the clusters created using an anomalous video, resulting in an improved discrimination between normal and abnormal events.
  \item Our framework achieves frame level AUC performance of 83.03\% on UCF-Crime \red{\cite{sultani2018real}} dataset and 89.67\% on ShanghaiTech \cite{shanghaiTech2017} dataset, outperforming the existing state-of-the-art approaches \red{\cite{zhong2019graph,lu2013abnormal,hasan2016anomaly,sultani2018real}.}
\end{itemize}
\vspace{-2mm}
\section{Related Work}
\noindent\textbf{Anomaly Detection as One Class Classification} Conventionally, anomaly detection has been tackled as learning normalcy in which test data is matched against learned representations of a normal class and the deviating instances are declared as anomalies.
Some researchers proposed to train one-class classifiers using handpicked features \cite{medioni2001event_twostream34,basharat2008learning_realworld7,wang2014learning_realworld38,zhang2009learning_twostream53,piciarelli2008trajectory_twostream36}
while others proposed to use the features extracted from pre-trained deep convolution models \cite{smeureanu2017deep,ravanbakhsh2017abnormal}.
Image regeneration based architectures \cite{Gong_2019_ICCV,ren2015unsupervised,xu2015learning_denoise,ionescu2019objectcentric,Nguyen_2019_ICCV,nguyen2019hybrid,xu2017detecting_denoise,sabokrou2017deep_novelty,sabokrou2018ALOCC}
make use of a generative network to learn normal data representations in an unsupervised fashion.
The intuition is that a generator cannot reconstruct unknown classes, hence it may generate high reconstruction errors while reconstructing anomalies. 
Few researchers also proposed pseudo-supervised methods in which fake anomaly examples are created using the normal data to transform the one-class problem into a binary-class problem \cite{ionescu2019objectcentric,zaheer2020old}. Our architecture, however significantly deviates from the one-class training protocol used in these methods as we utilize both normal and weakly-labeled anomalous data during training.

\noindent\textbf{Anomaly Detection as Weakly Supervised Learning}
This category utilizes the noisy annotations to carry out training on image datasets \cite{li2017learning,goldberger2016training,vahdat2017toward,patrini2017making,azadi2015auxiliary,natarajan2013learning,larsen1998design}. In such models, either the loss correction is applied \cite{azadi2015auxiliary} or noise models are specifically trained to separate out the noisy labeled data \cite{li2017learning,vahdat2017toward}. Our work is different from these image based weakly supervised methods as we try to tackle video based anomalies in which temporally-ordered sequence of frames are required.

In essence, the most similar works to ours are \cite{sultani2018real} and \cite{zhong2019graph}, which also attempt to train anomaly detection models using video-level annotations. Sultani et al. \cite{sultani2018real} propose to formulate the weakly supervised problem as Multiple Instance Learning (MIL). A video is considered as a bag of segments. To train the network, a ranking loss is computed between the top scoring segments from a normal and an anomalous bag. Each training iteration involves several pairs of such bags. In Zhong et al. \cite{zhong2019graph}, the authors attempt to clean noisy labels in anomalous videos by using graph convolutional neural networks. A training iteration is carried out based on one complete video from the training dataset. In contrast to these methods, our approach attempts to train a batch based model which learns to maximize scores of the anomalous parts of an input, where a batch corresponds to a portion of a training video. Furthermore, we also propose a normalcy suppression module which, by exploiting the \textit{noise free} labels in the normal labeled videos, learns to suppress the features corresponding to normal regions of a video. Lastly, a clustering distance based loss is also introduced which  improves  the  capability  of  our  model  to  produce better  anomaly  representations.

\noindent\textbf{Normalcy Suppression} The normalcy suppression module in our architecture can be seen as a variant of attention \astrid{\cite{vaswani2017selfattention,chen2018attention,shen2018egocentric,woo2018cbam,hu2018squeeze,wang2017residual}}. However, attributed to the rare occurrence of anomalies, we tackle the problem in terms of suppressing features as opposed to highlighting \astrid{\cite{woo2018cbam,hu2018squeeze,wang2017residual}}. Specifically, we define the problem by relying on the availability of \textit{noise free} normal video annotations. Therefore, unlike the conventional attention which utilizes a weighted linear combination of features as in attention, our model learns to suppress the normalcy by obtaining an element-wise product of the suppression scores with the features.

\begin{figure}[t]
\begin{center}
   \includegraphics[width=1\linewidth]{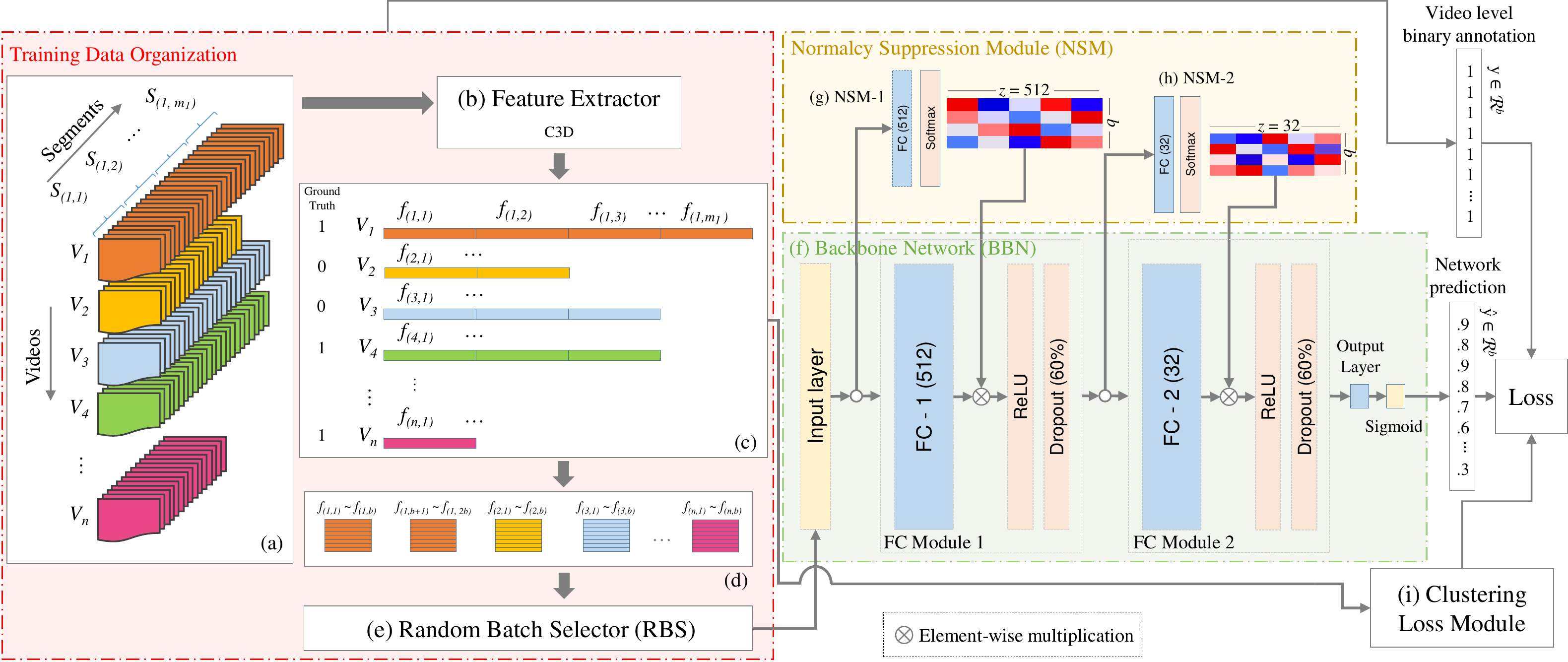}
\end{center}
   \caption{CLAWS Net: the proposed framework for anomaly detection using video-level weak supervision. (a) Input videos are divided into segments of equal length. (b)\&(c)  A feature vector is extracted from each segment. (d) Feature vectors of each video are divided into batches of the same size. (e) Batches are randomly selected for training. (f) Backbone network. (g)\&(h) Normalcy suppression modules. (i) Clustering loss module.
   }
\label{fig:architecture}
\vspace{-5mm}
\end{figure}
\section{Proposed Architecture}
\vspace{-3mm}
In this section, we present our CLAWS Net framework.
Various components of the proposed architecture, as shown in Fig. \ref{fig:architecture}, are discussed below:
\vspace{-2mm}
\subsection{Training Data Organization}
The proposed training data organization is shown in Fig. \ref{fig:architecture}(a)-(e). \astrid{Given $n$ training videos in a dataset,} each training video $V_i$ is divided into \astrid{$m_i$} non-overlapping segments $S_{(i,j)}$ of size $p$ frames, where $i \in [1, n]$ is the video index 
and \astrid{$j \in [1, m_i]$ is the segment index}. For each segment $S_{(i,j)}$, a feature vector $f_{(i,j)} \in \mathcal{R}^d$ is computed as $f_{(i,j)}$=$\mathcal{E}(S_{(i, j)})$ using a feature extractor $\mathcal{E}(\cdot)$ such as Convolution 3D (C3D) \cite{tran2015c3d}.
Consecutive feature vectors are arranged in batches \astrid{$B_k$, each consisting of $b$ feature vectors such that $B_k=\{f_{(i,j)}, f_{(i,j+1)}, \cdots, f_{(i,j+b-1)}\} \in \mathcal{R}^{b \times d}$, where $k \in [1, K]$ is the batch index of $K$ number of batches in the training data} which is used by the Random Batch Selector (RBS) in Fig. \ref{fig:architecture}(e) to retrieve batches randomly for training. All feature vectors maintain their temporal order within a batch, as shown in Fig. \ref{fig:architecture}(d). For each video we have binary labels as \{normal, abnormal\} represented as \{0, 1\}. As the training is performed in a weakly supervised fashion, each batch also inherits the labels of its features from the parent video.

In the existing weakly supervised anomaly detection systems, a training iteration (i.e. one weight update) is carried out on one or several complete videos \cite{sultani2018real,zhong2019graph}. In contrast to this practice, we propose several batches to be extracted from videos and then input to the backbone network in an arbitrary order using RBS. The configuration serves two purposes: 1) It minimizes correlation between consecutive batches while keeping the temporal order within batches which is necessary to carry out the weakly supervised anomaly detection training. 2) It allows our network to have more learning instances as our training iteration is carried out using a small portion of a video (batch) instead of a complete video. We also observed that breaking the temporal order between consecutive batches results in a significant increase in the backbone performance, which is discussed in the results section. 

\subsection{Backbone Network}

The backbone network (BBN) consists of two fully connected (FC) modules each containing an FC layer followed by a ReLU activation function and a dropout layer. The input layer receives each batch from RBS. 
The output layer has a sigmoid activation function \astrid{which produces anomaly score prediction $y \in \mathcal{R}^b$ of range $[0, 1]$}.
Training of the BBN is carried out using video-level labels. Hence, a batch from a normal video will have \astrid{labels $y = \mathbf{0} \in \mathcal{R}^b$ whereas a batch from an anomalous video will have labels $y = \mathbf{1} \in \mathcal{R}^b$, where $\mathbf{0}$ is an all-zeros vector, $\mathbf{1}$ is an all-ones vector, and $b$ is batch size.}

\begin{figure}[t]
\begin{center}
   \includegraphics[width=0.95\linewidth]{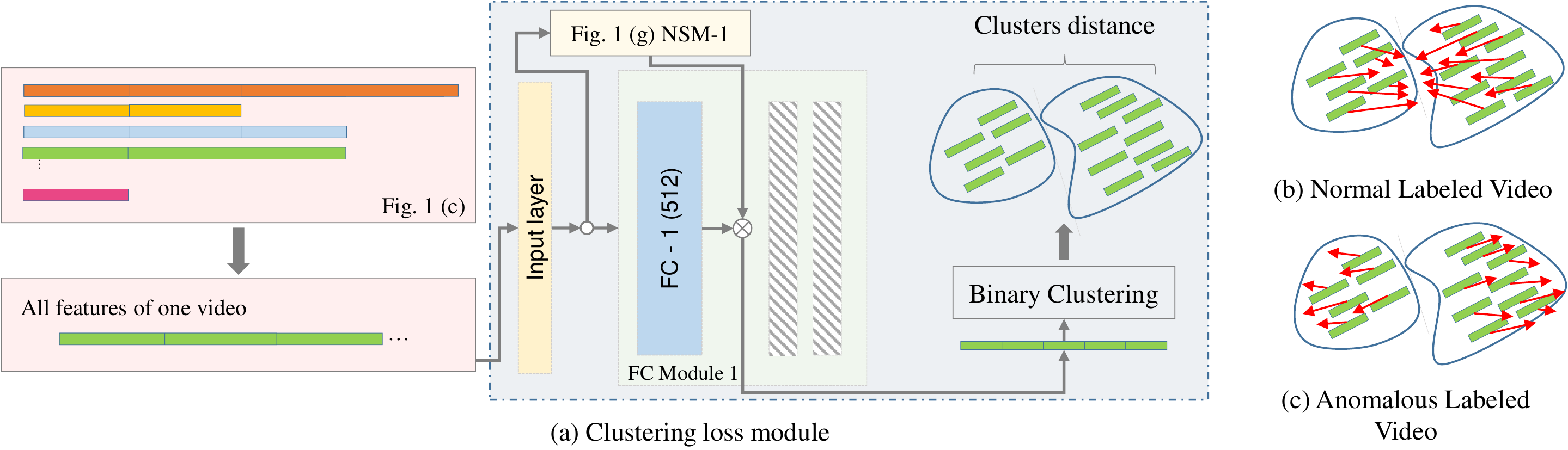}
   \vspace{-5mm}
\end{center}
   \caption{The proposed clustering loss module (a) which encourages our network to bring the clusters closer in the case of normal labeled videos (b) or push farther in the case of anomalous labeled videos (c). The FC Module 1 and the NSM-1 are same as the ones in Fig. \ref{fig:architecture} and are only used for inference in this module.}
\label{fig:CLM}
\vspace{-5mm}
\end{figure}

\subsection{Normalcy Suppression}
Our proposed architecture has multiple normalcy suppression modules (NSM) as shown in Figs. \ref{fig:architecture} (g)-(h). Each NSM contains an FC layer, kept consistent in dimensions with the FC layer in its respective FC Module, followed by a softmax layer. 
An NSM computes probability values across temporal dimension of the input batch therefore, serves as a global information collector over the whole batch.
Based on the FC layer dimension $z$ and the number of features $b$ in each input batch (Fig. \ref{fig:architecture} (d)), an NSM outputs a probability matrix $\mathcal{P}$ of size $b \times z$, such that the sum of each column in this matrix is 1. An element-wise multiplication between $\mathcal{P}$ and the output of FC layer is performed in the corresponding FC Module.

Our approach exploits the fact that all normal labeled videos have noise free labels at segment level as the anomalies do not appear in these videos.
During training, in case an input batch is taken from a normal labeled video, all features in this batch are labeled as normal and the BBN is expected to produce low anomaly scores on all input feature vectors. Therefore, the NSM learns to complement the BBN towards minimizing the overall training loss by distributing its probabilities across the whole input batch and not highlighting any portions of an input.
This phenomenon is particularly the desideratum behind choosing such element-wise configuration which provides more freedom to the NSM towards minimizing its values at each feature dimension of the whole batch.
In case an input batch is taken from an abnormal video and consequently all features of the batch are labeled as abnormal, it may be expected that the BBN will produce high anomaly scores on all input features. 
However, with normal batches as part of the training data, the BBN is also concurrently trained to produce low anomaly scores on the normal segments. Therefore, to some extent, the BBN achieves the capability to distinguish between normal and anomalous segments of an anomalous video. NSM further complements this capability of the BBN by suppressing the features of normal segments.
Given an anomalous batch as input, to minimize the overall training loss, NSM thus learns to suppress the portions of the input batch that do not contribute strongly towards the anomaly scoring in the BBN, consequently highlighting anomalous portions.

\subsection{Clustering Loss Module}
A clustering distance based loss is formulated with an intuition to encourage our network to successfully group deep video features into normal and anomalous clusters.
As mentioned previously, each feature vector $f_{(i,j)}$ inherits its label of being normal or abnormal from the parent video. A normal label means all segments are normal. However an abnormal label does not mean all segments are anomalous rather, some segments are anomalous while the remaining are normal. We propose to cluster a sub-representation of all the feature vectors $f_{(i,j)}$ of each training video into two clusters. In case of a normal video, we try to bring the centers of the two cluster as close to each other as possible assuming that both clusters correspond to normal segments (Fig. \ref{fig:CLM}(b)). In case of a video with abnormal label, we try to push the centers of the two clusters away from each other assuming that one cluster should contain normal segments while the other should contain abnormal (Fig. \ref{fig:CLM}(c)). As the clustering is performed on an intermediate representation inferred from the BBN, this loss results in an improved capability of our network to represent anomalies and consequently an enhanced anomaly detection performance of the proposed model.
Specifically, the distance $d_i$ between centers of the two clusters for a video $V_i$ containing $m_i$ number of segments is given as:
\begin{align}
  d_i = \frac{1}{m_i} \lVert \mathbf{c_1 - c_2} \rVert_2,
\end{align}
where $\mathbf{c_1}$ and $\mathbf{c_2}$ are the centers of the two clusters. As the training videos may have varying length, longer videos will have more batches than shorter videos therefore, $m_i$ is used to normalize distance values across the training dataset.

At the beginning of each training epoch, for all feature vectors $f_{(i,j)}$ of a video $V_i$, intermediate representations are computed using the BBN as shown in Fig. \ref{fig:CLM}(a). These resulting vectors are then grouped into two clusters using k-means clustering \cite{lloyd1982least} and $d_i$ is computed. This $d_i$ is then used with the corresponding batches at each training iteration.

\subsection{Training Losses of the Proposed Algorithm}
Training of our model is carried out to minimize regression, clustering distance, temporal smoothness, and sparsity losses, as explained below:

\noindent\textbf{Regression Loss:} The proposed CLAWS Net mainly performs regression to minimize mean square error using the video labels directly assumed towards each feature of the input batch:
    $\mathcal{L}_{pred} = \frac{1}{b}\sum_{l=1}^{b} (y_l - \hat{y}_l)^2$,
    where $y_l$ and $\hat{y_l}$ denote $l$-th ground truth and $l$-th predicted values in a batch respectively, and $b$ is the batch size.

\noindent\textbf{Clustering Distance Loss:} 
Given clustering distance ($d_i$), the clustering distance loss $\mathcal{L}_{c}$ of a video $V_i$ is given as:

\begin{equation}
    \mathcal{L}_{c}=
    \begin{cases}
      min(\alpha, d_i), & \text{if}\ V_i \text{ is Normal} \\
     \frac{1}{ d_i}, & \text{if}\ V_i \text{ is Abnormal},
    \end{cases}
\end{equation} where $\alpha$ is an upper bound on the clustering distance loss which helps to make the training
more stable in the presence of much varied video data.

\noindent\textbf{Temporal Smoothness Loss:}
It is applied based on the fact that most events are temporally consistent. Our proposed architecture  maintains temporal order among the feature vectors in each input batch therefore, similar to \cite{sultani2018real}, we apply temporal smoothness loss ($\mathcal{L}_{ts}$) as:
  $\mathcal{L}_{ts} = \sum_{l=1}^{b-1} (\hat{y}_{l+1}-\hat{y}_{l})^2$,
where $\hat{y}_l$ is the $l$-th prediction in a batch of size $b$. 

\noindent\textbf{Sparsity Loss:}
The sparsity loss, proposed previously in \cite{sultani2018real}, exploits the fact that anomalous events occur rarely as compared to the normal events. Hence, cumulative anomaly score of a complete video should be comparatively small. We compute this loss on each batch during training as:
 $\mathcal{L}_{s} = \sum_{l=1}^{b} \hat{y}_l$,
where $\hat{y}_l$ is the $l$-th prediction in a batch of size $b$. 

\noindent\textbf{Overall Loss Function:} Finally, complete loss of the proposed network is computed as:  
\begin{align}
  \mathcal{L} = \lambda_1\mathcal{L}_{pred} + (1-\lambda_1)\mathcal{L}_{c} +\lambda_2(\mathcal{L}_{s} +\mathcal{L}_{ts}),
\end{align}
where $\lambda_1$ and $\lambda_2$ assign  relative importance to different loss parameters.

\section{Experiments}
\subsection{Datasets}
Experiments have been conducted on two different video anomaly detection datasets 
including UCF-Crime \cite{sultani2018real} and ShanghaiTech \cite{shanghaiTech2017}.

\noindent\textbf{UCF-Crime} is a large-scale complex dataset which spans over 128 hours of videos (resolution 240 $\times$ 320 pixels), captured through CCTV surveillance cameras and contains 13 different classes of real world anomalies \cite{sultani2018real}. 
Its training split contains 800 normal and 810 anomalous videos, while the test split has 150 normal and 140 anomalous videos. Videos labeled as normal do not contain any abnormal scenes whereas, videos labeled as abnormal contain anomalous as well as normal scenes. In the training split, video-level binary labels are provided which can only be used by weakly supervised algorithms. In the test split the labels are provided at the frame level to facilitate the evaluation of anomaly localization. 

\noindent\textbf{ShanghaiTech}  is a medium-scale dataset for abnormal event detection captured in a university campus with staged anomalous events. It contains 437 videos ($317,398$ frames of resolution $480 \times 856$ pixels) captured at 13 different locations with varying lighting conditions and camera angles. The original protocol of this dataset is to train one-class classifiers which means that the training dataset contains only normal videos. \red{In order to make it suitable for evaluating weakly supervised binary classification architectures, }Zhong et al. \cite{zhong2019graph} reorganized the dataset into a mix of normal and anomalous videos in both testing and training splits. Their training split contains 175 normal and 63 anomalous videos and the test split contains 155 normal and 44 anomalous videos. For fair comparison, we follow this protocol for the training and evaluation of our system.

\subsection{Evaluation Metric}
Following previous works \cite{lu2013abnormal,zhong2019graph,sultani2018real,hasan2016anomaly}, we use Area Under the Curve (AUC) of the Receiver Operating Characteristic (ROC) curve, calculated with respect to the frame-level ground truth annotations of the test videos, as our evaluation metric. A larger AUC implies better discrimination performance at various thresholds. Since the number of normal frames in the test data is much larger than the number of anomalous frames, Equal Error Rate (EER) may not be a suitable measure \cite{sultani2018real}.

\begin{figure}[t]
\begin{center}
\begin{minipage}{.5\linewidth}
   \includegraphics[width=0.9\linewidth]{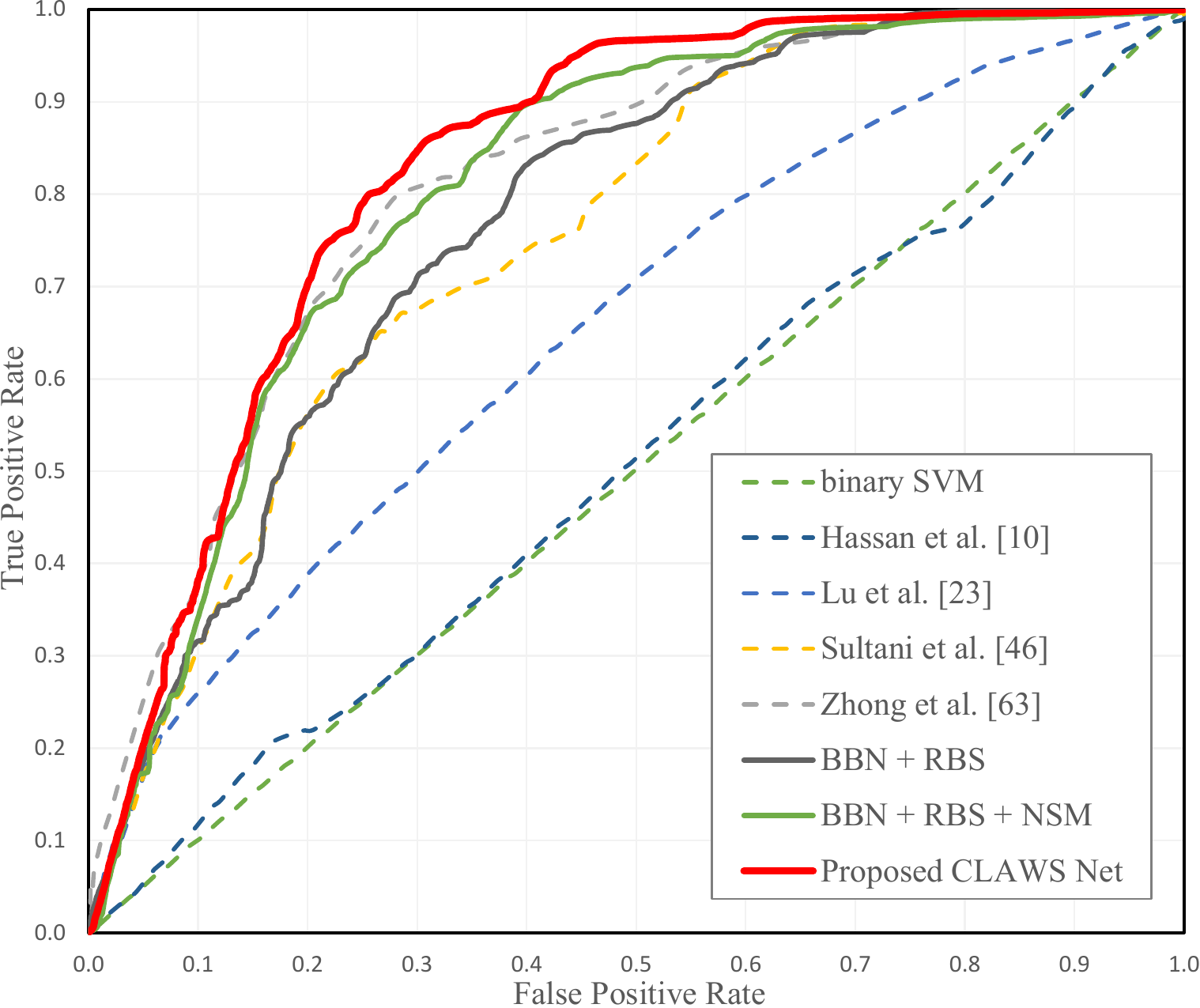}
   \captionsetup{width=.9\linewidth}
      \caption[width=0.5\linewidth]{ROC curves comparison with state-of-the-art using C3D features on UCF-Crime. }
\label{fig:ROC_plots}
\end{minipage}%
\begin{minipage}{.5\linewidth}
    \includegraphics[width=0.9\linewidth]{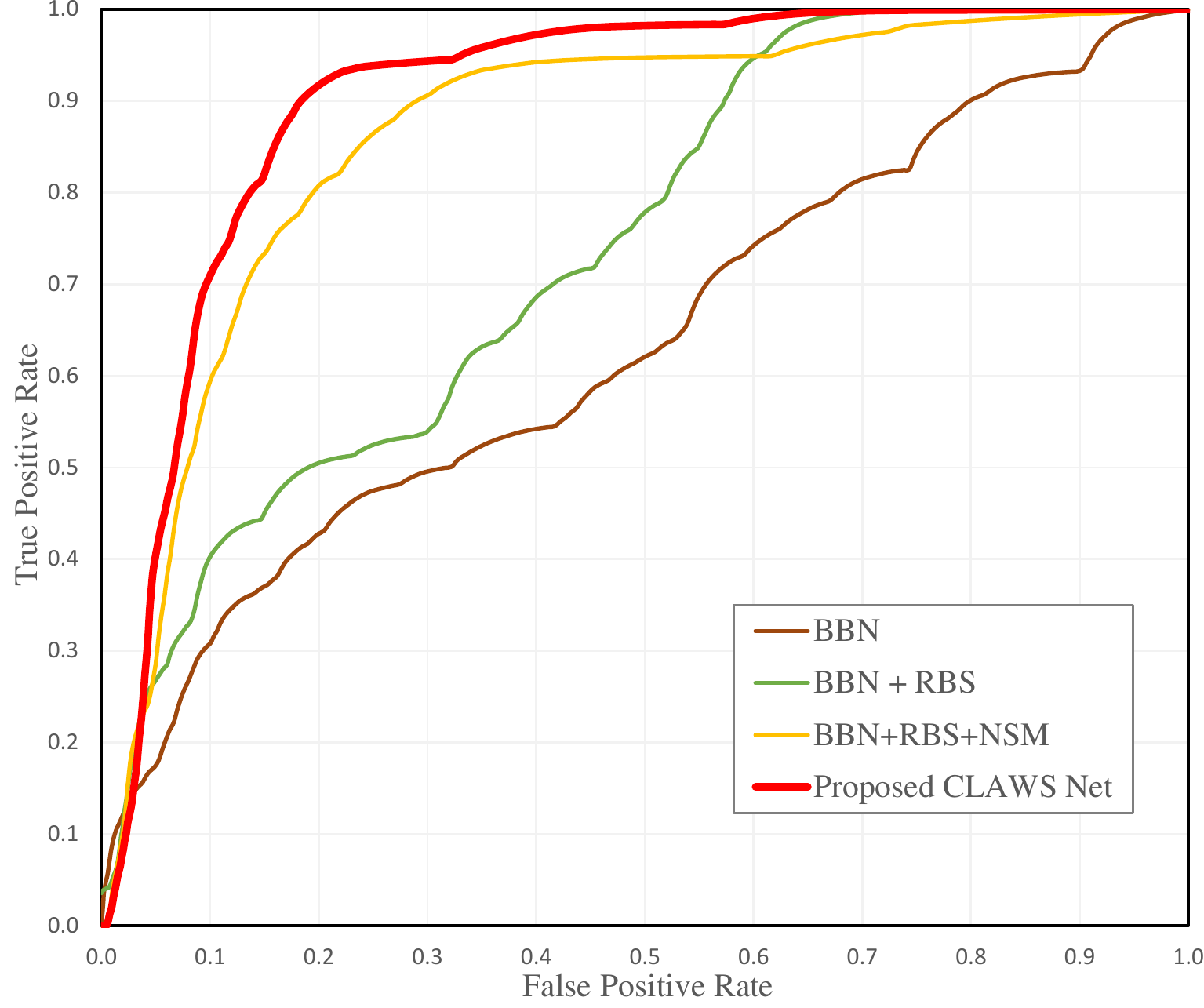}
    \captionsetup{width=.9\linewidth}
   \caption{ROC curves comparison of the variants of our proposed algorithm on ShanghaiTech.}
    \label{fig:shanghai_tech_roc}
\end{minipage}
\end{center}
\vspace{-6mm}
\end{figure}

\subsection{Experimental Settings}
Implementation of our model is carried out in PyTorch \cite{paszke2017pytorch}.  We train for a total of 100k iterations using RMSProp \cite{tieleman2012lectureRMSProp} optimizer with an initial learning rate of 0.0001 and a 10 times decrease after 80k iterations. In all our experiments, $\lambda_1$, $\lambda_2$ and $\alpha$ are set to $0.90$, $8.0\times10^{-5}$ and 1.0, respectively. FC layers of the FC Module 1 and FC Module 2 are set to have 512 and 32 channels. C3D \cite{tran2015c3d}, a 3D-convolution model for videos pre-trained on Sports-1M dataset \cite{karpathy2014Sports1M}, is employed as the feature extractor in Fig. \ref{fig:architecture} (b). Default feature extraction settings of C3D, as proposed in \cite{tran2015c3d,zhong2019graph,sultani2018real}, are used. \red{Mean normalization is applied on the extracted features.} Batch size \red{$b$} is set to $64$ feature vectors, segment $S_{(i,j)}$ size is set to $16$ frames and feature $f_{(i,j)}$ size $d$ is 2048.


\begin{table}[t]
\begin{center}
\caption{Frame-level AUC performance comparison on UCF-Crime (BBN: Backbone network, RBS: Random batch Selector, NSM: Normalcy suppression module).
}
\begin{tabular}{c|c|c} 
\text{\textbf{Method}} & \text{\textbf{Features Type}} & \text{\textbf{AUC(\%)}} \\ \hline
SVM Baseline   & C3D          & 50.00\\ \hline
Hasan et al. \cite{hasan2016anomaly}    & C3D                      & 50.60\\ \hline
Lu et al. \cite{lu2013abnormal}     & C3D                     & 65.51\\ \hline
Sultani et al. \cite{sultani2018real}   & C3D                       & 75.41\\ \hline
Zaheer et al. \cite{zaheer2020cleaning}   & C3D                       & 78.27\\ \hline
SRF \cite{zaheer2020self}   & C3D                       & 79.54\\ \hline
GraphConv. et al. \cite{zhong2019graph} &    C3D  &  81.08\\ \hline
GraphConv. \cite{zhong2019graph} &    $TSN^{Optical Flow}$  &  78.08\\ \hline
GraphConv. \cite{zhong2019graph} &    $TSN^{RGB}$  &  82.12\\ \hline\hline
BBN  &    C3D &  69.50\\ \hline
BBN + RBS  &    C3D &  75.95\\ \hline
BBN + RBS + NSM  &    C3D &  80.94\\ \hline
\textbf{Proposed CLAWS Net} &    \textbf{C3D}& \textbf{83.03}
\end{tabular}
\label{tab:UCF_crime_AUC}
\end{center}
\vspace{-5mm}
\end{table}
\vspace{-2mm}

\subsection{Experiments on UCF-Crime Dataset}
\vspace{-1mm}
We train our proposed model on UCF-Crime dataset using only video-level labels. \red{Fig. \ref{fig:ROC_plots}} and Table \ref{tab:UCF_crime_AUC} visualize a comparison of our method with current state-of-the-art approaches. In both types of comparisons, our CLAWS Net shows superior performance than the compared algorithms. Interestingly, our proposed RBS enhances the performance of the BBN to 75.95\% which is superior than the performance reported by Hasan et al. \cite{hasan2016anomaly}, Lu et al. \cite{lu2013abnormal} and Sultani et al. \cite{sultani2018real}.
It is because of the significant decrease in the inter-batch correlation which improves the learning of BBN. 
Noticeably, compared to Zhong et al. \cite{zhong2019graph}, our approach is fairly simple as we do not train a deep action classifier. However, using the similar C3D features, our CLAWS Net depicts 1.95\% improved performance.

\subsection{Experiments on ShanghaiTech}
The CLAWS Net framework is evaluated on ShanghaiTech following the test/train split proposed by Zhong et al. \cite{zhong2019graph}. Being a recent split, other existing methods have not reported performance on this dataset. Using the same protocol, our framework outperforms Zhong et al. \cite{zhong2019graph} by a significant margin of 13.23\% when both algorithms use similar C3D features (see Table \ref{tab:shanghaitechAUC}). Zhong et al \cite{zhong2019graph} reported relatively better performance using TSN features, however our proposed model outperforms their TSN based performance as well by 5.23\%. In our work, the reported results are evaluated using C3D features which is in consistence with most of the existing methods. However, using features with better representation may result in a further improved performance. An ROC curve based performance comparison of the three variants of our proposed approach is provided in Figure \ref{fig:shanghai_tech_roc}. 
We observe a performance boost of 12.14\% by adding RBS to the BBN. Furthermore, \red{8.12\%} boost is observed by using NSM along with BBN+RBS. Finally the complete system, which incorporates all losses as well, further enhances the performance by \red{1.91\%}. These experiments highlight the importance of each component in CLAWS Net.  


\begin{table}[t]
\begin{center}
\caption{Frame-level AUC performance comparison on ShanghaiTech (BBN: Backbone network, RBS: Random batch Selector, NSM: Normalcy suppression module).}
\begin{tabular}{c|c|c}
\textbf{Method} & \text{\textbf{Features Type}}        & \textbf{AUC \%} \\ \hline
GraphConv. \cite{zhong2019graph} &C3D & 76.44  \\ \hline
GraphConv. \cite{zhong2019graph} &$TSN^{Optical Flow}$ & 84.13  \\ \hline
GraphConv. \cite{zhong2019graph} &$TSN^{RGB}$ & 84.44 \\ \hline
Zaheer et al. \cite{zaheer2020cleaning}   & C3D                       & 84.16\\ \hline
SRF \cite{zaheer2020self}   & C3D                       & 84.16\\ \hline\hline
BBN  &    C3D &  67.50\\ \hline
BBN + RBS  &    C3D &  79.64\\ \hline
BBN + RBS + NSM  &    C3D &  87.76\\ \hline
\textbf{Proposed CLAWS Net} &    \textbf{C3D} & \textbf{89.67}
\end{tabular}
\label{tab:shanghaitechAUC}
\end{center}
\vspace{-3mm}
\end{table}

\begin{table}[t]
\begin{center}
\caption{Ablation studies of our proposed approach on UCF-Crime (BBN: Backbone network, RBS: Random batch Selector, NSM: Normalcy suppression module, $\mathcal{L}_s$: Sparsity loss, $\mathcal{L}_{tc}$: Temporal consistency loss, $\mathcal{L}_c$: Clustering distance loss).}
\resizebox{0.7\linewidth}{!}{
\begin{tabular}{|>{\centering}p{13mm}|>{\centering}p{13mm}|>{\centering}p{13mm}|>{\centering}p{13mm}|>{\centering}p{13mm}|>{\centering}p{13mm}||c|}
\hline
\multicolumn{1}{|c|}{\textbf{BBN}} & \multicolumn{1}{c|}{\textbf{RBS}} & \multicolumn{1}{c|}{\textbf{NSM-1}} & \multicolumn{1}{c|}{\textbf{NSM-2}} & \multicolumn{1}{c|}{\textbf{$\mathcal{L}_s + \mathcal{L}_{tc}$}} & \multicolumn{1}{c||}{\textbf{$\mathcal{L}_c$}} & \multicolumn{1}{c|}{\textbf{AUC (\%)}} \\ \hline
\multicolumn{7}{|c|}{Bottom-up Approach} \\ \hline
\ding{51} & - & - & - & - & - & 69.50 \\ \hline
\ding{51} & \ding{51} & - & - & - & - & 75.95 \\ \hline
\ding{51} & \ding{51} & \ding{51} & - & - & - & 78.60 \\ \hline
\ding{51} & \ding{51} & \ding{51} & \ding{51} & - & - & 80.94 \\ \hline
\ding{51} & \ding{51} & \ding{51} & \ding{51} & \ding{51} & - & 81.53 \\ \hline
\multicolumn{7}{|c|}{Top-down Approach} \\ \hline
\ding{51} & - & \ding{51} & \ding{51} & \ding{51} & \ding{51} & 80.23 \\ \hline
\ding{51} & \ding{51} & - & \ding{51} & \ding{51} & \ding{51} & 77.39 \\ \hline
\ding{51} & \ding{51} & \ding{51} & - & \ding{51} & \ding{51} & 79.78 \\ \hline
\ding{51} & \ding{51} & - & - & \ding{51} & \ding{51} & 76.81 \\ \hline
\ding{51} & \ding{51} & \ding{51} & \ding{51} & - & \ding{51} & 82.41 \\ \hline\hline
\ding{51} & \ding{51} & \ding{51} & \ding{51} & \ding{51} & \ding{51} & 83.03 \\ \hline
\end{tabular}
}
\label{tab:ablation}
\end{center}
\vspace{-7mm}
\end{table}

\subsection{Ablation}
We performed two types of ablation studies by using bottom-up as well as top-down approaches on the UCF-Crime dataset as shown in Table \ref{tab:ablation}. In the former approach, we started with the evaluation of only Backbone  Network (BBN) and kept on adding different modules while observing the performance boost. In the later approach we started with the whole CLAWS Net and removed different modules to observe the consequent performance degradation.   

\begin{figure}[t]
\begin{center}
  \includegraphics[width=1.0\linewidth]{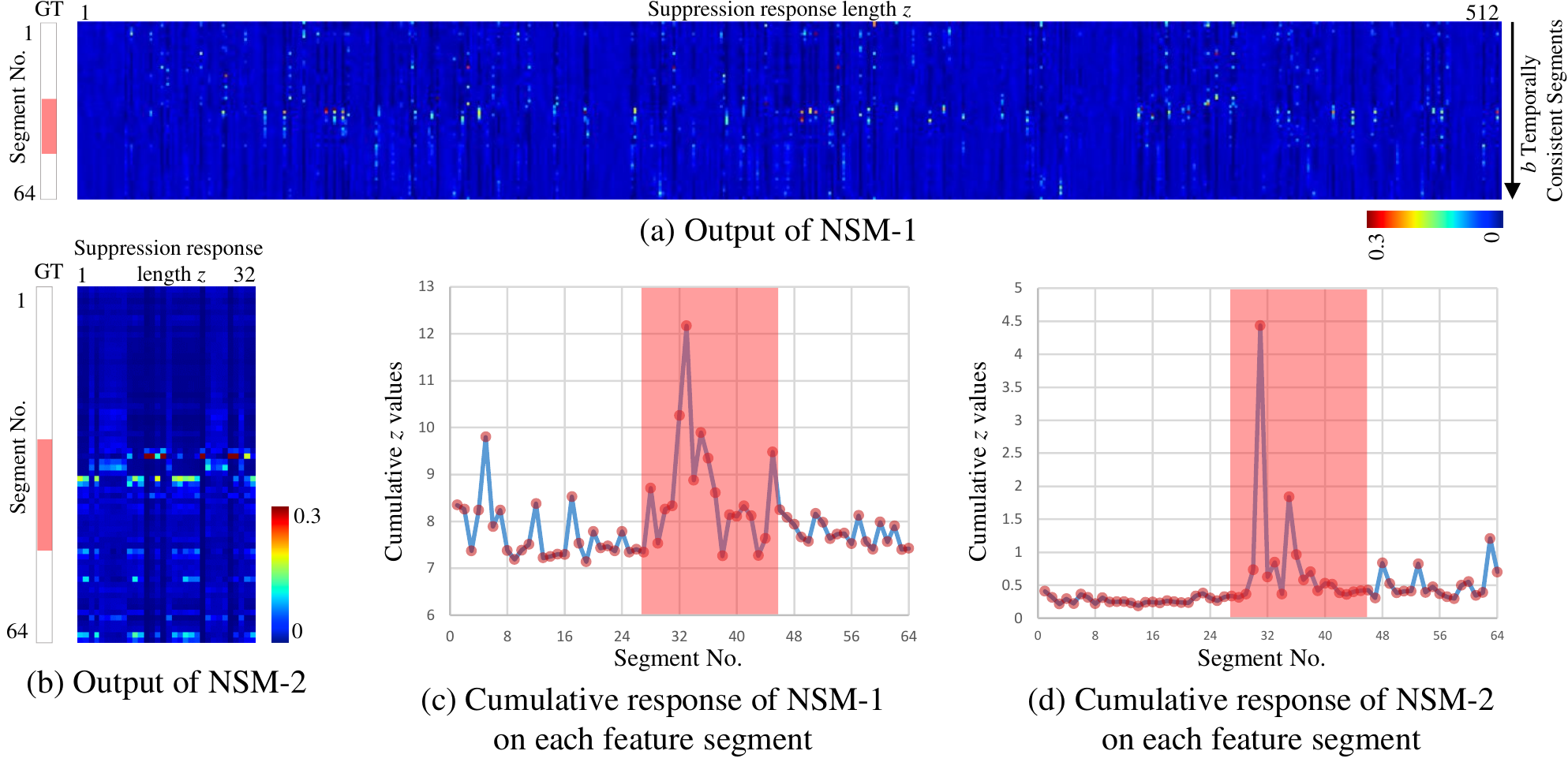}
\end{center}
  \caption{Softmax output of our proposed NSM-1 (a) and NSM-2 (b) on an input batch from an anomalous test video. It can be seen that both modules learn to successfully suppress the normal regions of an input. Cumulative suppression values of these modules on each individual feature of segments is visualized in (c) \& (d) which also provides an overall insight on the response of each NSM towards an anomalous input. Red colored windows show anomaly ground truth.}
\label{fig:attention_batches}
\vspace{-5mm}
\end{figure}

The BBN achieves 69.5\% AUC while the addition of RBS enhances the performance to 75.95\% which validates its importance in our approach.
Addition of the NSM-1 results in an improved performance of 78.60\% whereas the NSM-2 further elevates the performance to 80.94\%. The overall boost in the AUC by NSM-1 is larger than by NSM-2. It can mainly be attributed to the size of the FC layer in NSM-1, which is 16 times larger than the FC layer in NSM-2. 
Figs. \ref{fig:attention_batches} (a) \& (b) show the responses of NSM-1 \& 2 for an anomalous test video batch. The response corresponding to the anomalous events is much higher compared to the response on normal events. These values are multiplied with the FC layer output and thus suppress the values corresponding to the normal events.   
Figs. \ref{fig:attention_batches} (c) \& (d) show the cumulative response of NSM-1 and NSM-2 modules computed by summation along the response length $z$. The cumulative response in the anomalous region is significantly higher than the response in the normal regions. It clearly demonstrates that the NSM modules successfully learn to suppress the normal regions, consequently highlighting the anomalous regions, at two intermediate levels of the network.
Table \ref{tab:ablation} also shows that the addition of temporal consistency and sparsity losses brings the performance to 81.53\%, and finally the addition of the clustering distance loss yields an overall system performance of 83.03\%. This study depicts the importance of each component in our proposed architecture.

In the top-down approach (Table \ref{tab:ablation}) deletion of RBS from our CLAWS Net causes a drop of 2.8\% AUC. Compared to this, the addition of RBS to BBN in the bottom-up approach resulted in an improvement of 6.45\%. Thus, some of the performance of RBS is compensated by the other components in the complete system, however it cannot be fully replaced. Furthermore, deletion of NSM-1 and NSM-2 from the overall architecture causes a drop of 5.64\% and 3.25\% respectively. Consistent to the results in the bottom-up approach, it demonstrates a relatively higher importance of NSM-1 than NSM-2, mainly due to the larger size of its FC layer.
Another contributing factor is that the NSM-1 directly affects the clustering loss module. As shown in Fig. \ref{fig:CLM}, the clustering is performed after multiplying the NSM-1 response with the output of FC-1. If both NSM-1 and NSM-2 are removed, the remaining system achieves 76.81\% AUC which is 6.22\% less than the complete CLAWS Net performance. However, the addition of NSM-1 and NSM-2 to BBN+RBS in the bottom-up approach caused an improvement of 4.99\%. This demonstrates more importance of the NSM modules than what is observed in the bottom-up approach due to their direct effect on the clustering loss module as well as indirect effect on other losses. 

\begin{figure}[t]
\begin{center}
   \includegraphics[width=1\linewidth]{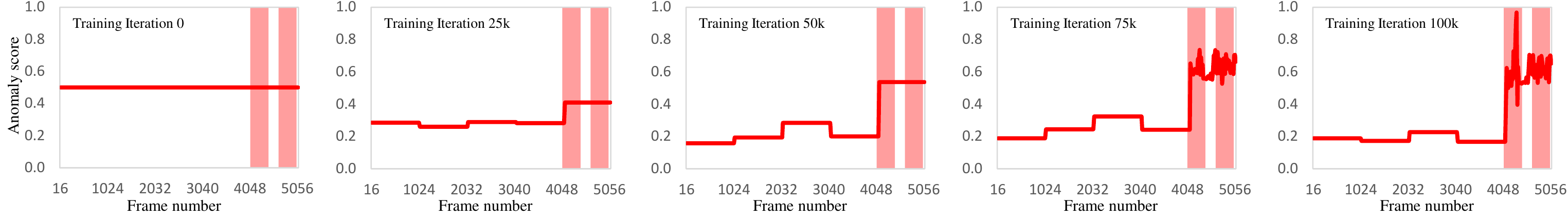}
\end{center}
\vspace{-2mm}
   \caption{Evolution of frame-level anomaly scores output by our network over several training iterations. Although weakly supervised, our framework learns to produce significantly higher scores in the anomalous portions whereas low scores in the normal portions of a video. Red colored windows show anomaly ground truth.}
\label{fig:scores_evolution}
\vspace{-2mm}
\end{figure}

\vspace{-2mm}
\subsection{Qualitative Analysis}
\begin{figure}[t]
\begin{center}
   \includegraphics[width=1\linewidth]{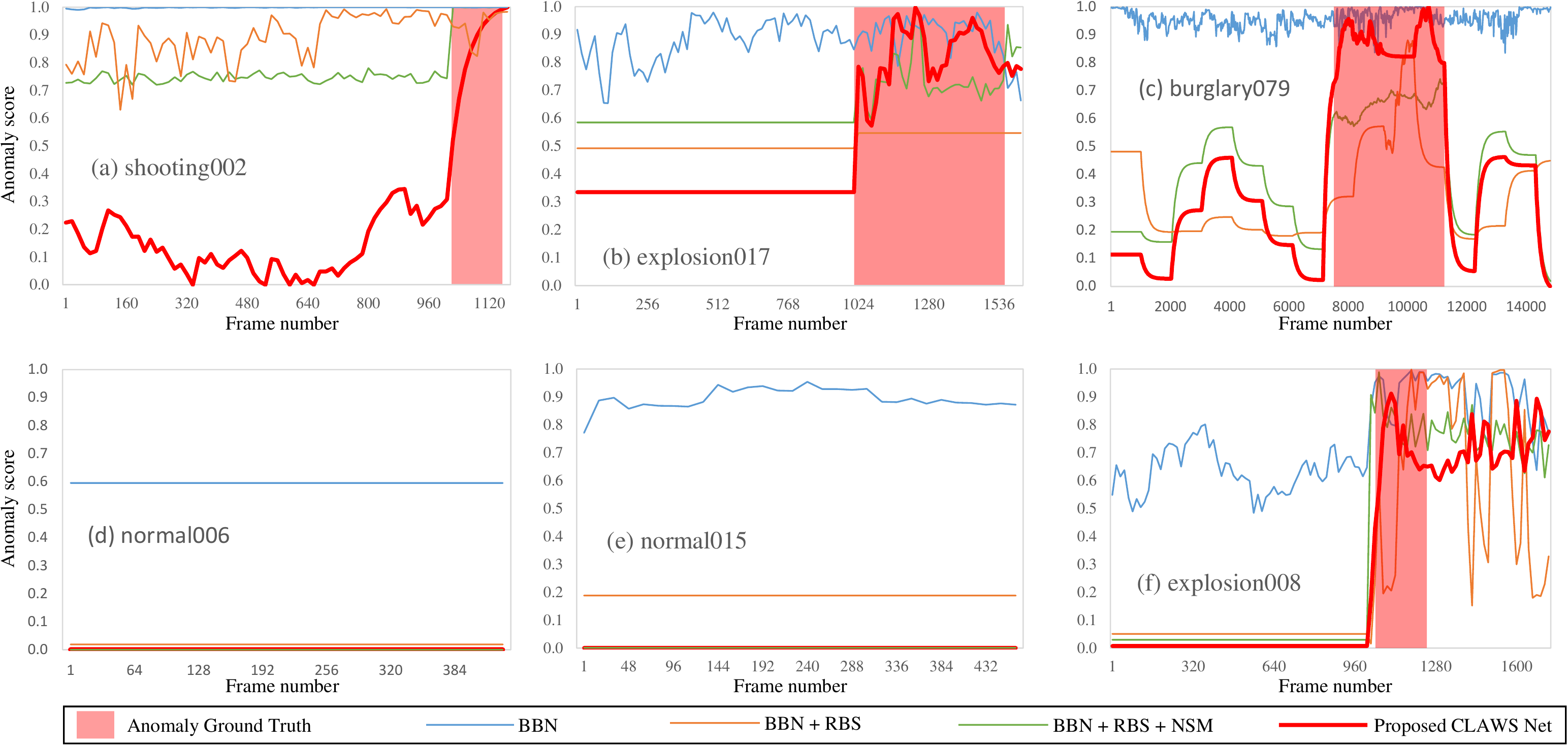}
\end{center}
\vspace{-2mm}
   \caption{Qualitative results of our method on test videos of the UCF-Crime dataset. (a), (b) \& (c) show anomaly cases while (d) \& (e) depict normal cases. (f) illustrate a relatively unsuccessful case of our system in which, the network keeps showing higher anomaly scores after the explosion. Red colored windows show anomaly ground truth.}
\label{fig:qualitative_results}
\vspace{-2mm}
\end{figure}

The evolution of the proposed anomaly detection system over several training iterations is shown in Fig. \ref{fig:scores_evolution}. As the number of training iterations increases, the difference between the response of our system on normal and anomalous regions also increases. 
Fig. \ref{fig:qualitative_results} provides a comparison of the anomaly scores produced by different configurations of our proposed architecture on two normal and four anomalous test videos from the UCF crime dataset. In these cases, the BBN was not able to accurately discriminate between the normal and anomalous regions. The addition of RBS showed significant improvements except in the $shooting$\textit{002} case where the improvement is small. 
It can be observed that with the addition of normalcy suppression modules, the difference between anomaly scores on normal and abnormal regions became higher which is a desirable property in the anomaly detection systems. The proposed suppression not only pushed the anomaly scores of the normal regions towards 0 but also created a smoothing effect. 
The response of our complete system, CLAWS Net, is more stable as well as discriminative than all the other variants in most of the cases. It should also be noted that Fig. \ref{fig:qualitative_results} (f) shows a relatively unsuccessful case in which the system continues to show high anomaly score even after the anomalous event is over, which is not unlikely due to the aftermath of an explosion. The annotations in the dataset are marked only for the duration of explosion, while the scenes in such an event may remain abnormal for a significantly longer time. 
\vspace{-3mm}

\section{Conclusions}
\vspace{-2mm}
In this study, we present a weakly supervised anomalous event detection system trainable using only video-level labels.
Unlike the existing systems which utilize complete video based training iterations, we adopt a batch based training. A batch may have several temporally ordered segments of a video and one video may be divided into several batches depending on its length. Selection of these batches in random order helps breaking inter-batch correlation and demonstrates a significant boost in performance.
A normalcy suppression mechanism is also proposed which collaborates with the backbone network in detecting anomalies by learning to suppress the features corresponding to the normal portions of an input video. Moreover, a clustering distance based loss is formulated, which improves the capability of our network to better represent the anomalous and normal events.
Validation of the proposed architecture on two large datasets (UCF-Crime \& ShanghaiTech) demonstrates SOTA results.

\vspace{-2mm}
\section{Acknowledgements}
\vspace{-2mm}
This work was supported by the ICT R\&D program of MSIP/IITP. [2017-0-00306, Development of Multimodal Sensor-based Intelligent Systems for Outdoor Surveillance Robots]

\clearpage
%
%
\bibliographystyle{splncs04}
\bibliography{4066}
\end{document}